\documentclass{article}
\usepackage{spconf,graphicx}
\usepackage{amsmath}
\usepackage{amssymb}
\usepackage[table]{xcolor}%
\usepackage[mathscr]{euscript}
\usepackage{xcolor}
\usepackage[normalem]{ulem}
\usepackage{url}
\usepackage{multirow}
\usepackage{graphicx}
\usepackage{hyperref}

\title{{Highly Constrained Coded Aperture Imaging Systems Design Via a Knowledge Distillation Approach}}
\name{Leon Suarez-Rodriguez, Roman Jacome, Henry Arguello\thanks{This work was funded by the Vicerrector\'ia de Investigaci\'on y Extensi\'on from Universidad Industrial de Santander under Project 3925.}}
\address{Universidad Industrial de Santander}
\begin{document}
%
\maketitle
\begin{abstract}
 Computational optical imaging (COI) systems have enabled the acquisition of high-dimensional signals through optical coding elements (OCEs). OCEs encode the high-dimensional signal in one or more snapshots, which are subsequently decoded using computational algorithms. Currently, COI systems are optimized through an end-to-end (E2E) approach, where the OCEs are modeled as a layer of a neural network and the remaining layers perform a specific imaging task. However, the performance of COI systems optimized through E2E is limited by the physical constraints imposed by these systems. This paper proposes a knowledge distillation (KD) framework for the design of highly physically constrained COI systems. This approach employs the KD methodology, which consists of a teacher-student relationship, where a high-performance, unconstrained COI system (the teacher), guides the optimization of a physically constrained system (the student) characterized by a limited number of snapshots. We validate the proposed approach, using a binary coded apertures  single pixel camera for monochromatic and multispectral image reconstruction. Simulation results demonstrate the superiority of the KD scheme over traditional E2E optimization for the designing of highly physically constrained COI systems. 
\end{abstract}



\section{Introduction}
\label{sec:intro}









 Computational optical imaging (COI) integrates optical systems with computational algorithms to overcome traditional imaging limitations, including dynamic range, spatial resolution, and depth of field constraints \cite{bhandari2022computational}. COI has enabled the acquisition of high-dimensional signals, demonstrating excellent performance in applications such as medical imaging, smartphone photography, and autonomous driving \cite{bhandari2022computational}. {However, current measurement optical} devices are constrained to acquiring low-dimensional intensity values of high-dimensional scenes. To address this, COI systems employ optical coding elements (OCEs) to encode high-dimensional scenes in one or more snapshots, {which are subsequently decoded} using computational algorithms.  The effectiveness of these systems relies on the design of the pattern of OCEs and the computational approach employed for tasks, such as spectral image reconstruction, pose estimation, and image classification \cite{E2E_PROF_HENRY}.

{Coded apertures (CAs) are commonly used as a type of OCE with various} applications, including depth estimation, X and gamma-ray detection, hyperspectral image reconstruction, and super-resolution \cite{CAS_REVIEW, CAS_DEPTH, CAS_PROF_HENRY}. Particularly, the single-pixel camera (SPC) is one of the most important COI systems based on CAs {\cite{SPC_COMPRESSIVE_SAMPLING}}. CAs form an arrangement of elements that modulate the incoming wavefront, they can be either real-valued or binary \cite{CAS_PROF_HENRY, SPC_JORGE}. Real valued CA elements attenuate the wavefront at different levels allowing a wide range of admitted values but they are harder to fabricate, and calibrate, {leading to} more storage space, and slower processing speed than binary CAs {\cite{E2E_JORGE_IEEE}}. {Binary CAs, contain opaque and translucent elements that either let pass the light or block it at each spatial point \cite{SPC_HANS}. In particular}, binary CAs are the preferred choice in practical applications  \cite{E2E_JORGE_IEEE, E2E_JORGE_OPTICA}. To address practical challenges and enhance acquisition speed in COI systems, OCEs may also be constrained {according to} the number of snapshots \cite{E2E_JORGE_IEEE}.
The primary goal of capturing multiple snapshots is to enhance information acquisition, ultimately improving { the performance of a given imaging task such as recovery, classification, or segmentation \cite{E2E_PROF_HENRY}}. {Then}, a trade-off exists between task performance and the time required for acquiring and processing additional snapshots {is established in} \cite{E2E_JORGE_IEEE}. 

A key aspect of CA imaging lies in the optimal design of CAs, {such that the number of snapshots is minimized}. Currently, state-of-the-art CA design relies on deep learning methods utilizing end-to-end optimization techniques \cite{E2E_PROF_HENRY, E2E_JORGE_IEEE}. In this context, the {sensing model} is modeled as a layer of a neural network and the remaining layers perform any particular imaging task such as segmentation, object detection, {or} reconstruction. The neural network is trained simultaneously to learn the OCEs of the COI system and the parameters of the neural network for a given imaging task \cite{E2E_JORGE_IEEE}. {However, the performance of COI systems optimized through E2E optimization is limited, due to the heavily constrained optimization to represent the physical limitations of COI systems, such as the binary nature of the CAs, restrictions in the number of snapshots, and speed of acquisition. These issues harm the gradient computation during the network training leading to subpar optimization  \cite{SPC_ROMAN}.} {To overcome these limitations, this paper proposes a new learning strategy that leverages a knowledge distillation (KD) approach.}

Particularly, KD proposed by \cite{KD_HINTON}, is a model compression technique where a smaller model (student) is trained to replicate the behavior of a larger and more complex model (teacher). This process involves transferring knowledge from the teacher to the student model to reduce physical constraints such as storage space and computational requirements while preserving its performance \cite{KD_HINTON}. While KD has been extensively explored in high-level vision tasks like classification, segmentation, and detection, limited attention has been given to low-level tasks such as denoising and super-resolution \cite{KD_DENOISING, DISTILLATION_SUPER_RES_3}. {Additionally, in the computational imaging field, KD has been employed in magnetic resonance imaging reconstruction without considering the design of the sensing model~\cite{KD_MRI}} 



 {In this work, KD is applied to optimize the OCEs in highly physically constrained COI systems. Instead of optimizing the OCEs to allow a high-fidelity reconstruction of the original image, they are optimized to imitate the behavior of an unconstrained high-performance COI system.} {To validate the proposed approach for the design of OCEs, a SPC system is used for {monochromatic and spectral image reconstruction}}. KD allows the transfer of knowledge from a high-performance, low-constrained COI system to a highly physically constrained COI system. This process enables an increase in processing speed, and a reduction in storage space and the number of snapshots while maintaining or even improving reconstruction performance. The reconstruction network incorporates an unrolling network based on the ADMM algorithm \cite{ADMM,algorithmunrolling}. The proximal operator of the ADMM algorithm, which promotes sparsity of the signal, is learned through a neural network  \cite{ISTA_NET}. The first layer of the reconstruction network learns the optimal CAs, while the remaining layers learn the reconstruction of the re-projected measurements of the scene. A correlation loss is employed to transfer knowledge from the teacher to the student \cite{KD_LOSS_CORRELATION}. Specifically, this loss encourages a congruence correlation between the sparse representation of the teacher and student features at every stage of the ADMM unrolling algorithm. Furthermore, the Euclidean norm is employed to ensure that the output of the student matches the output of the teacher. Simulations demonstrate the superiority of the proposed approach over the traditional E2E deep learning scheme for designing highly physically constrained COI systems. 

\section{E2E Optimization} \label{E2E_optimization}

{Consider a high dimensional signal $\mathbf{x} \in \mathbb{R}^N$ that is acquired by a low-dimensional projected encoded measurements $\mathbf{y} \in \mathbb{R}^M$. The sensing procedure is modeled as a differentiable linear operator
\begin{equation}
\label{eq:sensing_model}
    \mathbf{y} = \mathbf{H_\Phi} \mathbf{x} + \boldsymbol{\omega},
\end{equation}
\noindent where $\mathbf{H_\Phi} \in \mathbb{R}^{M \times N}$ is the sensing matrix of the linear {sensing model}, $\boldsymbol{\Phi}$ are the OCEs of the sensing system and $\boldsymbol{\omega} \in \mathbb{R}^M$ is additive noise.
Traditional deep learning approaches for designing OCEs employ an E2E optimization framework. In this approach, the OCEs are modeled as a layer of a neural network, and the remaining layers of the network perform a specific imaging task. Subsequently, the OCEs $\mathbf{\Phi}$ and the parameters of the reconstruction network $\boldsymbol{{\theta}}$ are jointly learned following the optimization problem \cite{E2E_PROF_HENRY,SPC_JORGE} \vspace{-0.3cm}

{
\begin{equation}
\label{eq:e2e_optimization}
\begin{aligned}
 \{\boldsymbol{\theta^*}, \mathbf{\Phi^*}\} 
     =\underset{\substack{\mathbf{\Phi} \in \mathcal{E}, \boldsymbol{\theta} }}{\arg \min } \ \frac{1}{P} \sum_{p=1}^P \mathcal{L}_\text{task} & \left( \psi_{\boldsymbol{\theta}}(\mathbf{H_\Phi} \mathbf{x}_p, \mathbf{d}_p)
 \right) ,\\
\end{aligned}
\end{equation}}
\noindent where $\{\mathbf{x}_p,\mathbf{d}_p\}_{p=1}^{P}$ is the training dataset, $\mathbf{d}_p$ is the ground truth, and $\mathcal{L}_{\text{task}}$ is the loss function for a given imaging task, such as spectral reconstruction or image classification. The set $\mathcal{E}$ constrains the optimization and models the values of $\boldsymbol{\Phi}$ to promote specific properties, such as a reduced number of snapshots or binary CAs due to physical limitations \cite{CAS_PROF_HENRY}.
$\psi_{\boldsymbol{\theta}}$ depicts {a neural} network {to perform the imaging task}, and $\{\mathbf{\Phi^*}, \boldsymbol{\theta^*}\}$ represents the set of optimal OCEs and optimal parameters of the reconstruction network, respectively. {Equation \eqref{eq:e2e_optimization} is solved by stochastic gradient descent (SGD) algorithm}. Once the optimal OCEs $\mathbf{\Phi^*}$ for the given task are learned, {they are implemented in the physical optical sensing system to acquire real scenes.}  The learned network can subsequently be applied to perform the given task on the acquired scenes. 

 }

\section{Learning Constrained Optical Encoding via Knowledge Distillation} \label{UNROLLING}

To surpass the highly constrained optimization of OCE in the E2E optimization, we devise a KD approach wherein from a high-performance low-constrained optimized optical system, we distill its performance to a highly constrained optical system. To this end, we developed a KD framework based on an unrolling network \cite{algorithmunrolling}, denoted as $\psi_{\boldsymbol{\theta}}$,  to perform the distillation on the OCE design. {First, we derive the unrolling network formulation, then, we mathematically describe our KD approach.}

\subsection{Recovery from Unrolling Network}

{Unrolling Networks has been a widely employed approach for solving inverse problems due to the synergistic combination of traditional optimization formulations and deep learning methods, with remarkable results in broad inverse problems such as in image deblurring \cite{li2019algorithm}, compressive spectral imaging \cite{wang2019hyperspectral}, magnetic resonance imaging \cite{zhang2022high}. Here, a deep neural network is designed to perform iterations of an iterative recovery algorithm{,} e.g., ADMM  \cite{ADMM}. }
The unrolling network employed in this work is formulated inspired by the following optimization problem

\begin{equation}
\underset{\mathbf{x}}{\text{minimize }} \frac{1}{2}\|\mathbf{y}-\mathbf{H_\Phi x}\|_2^2+\lambda{R(\mathbf{x})}, 
\label{eq:optimization_problem}
\end{equation}

where the first term is the fidelity term that guarantees the reconstruction is consistent with the observation and the second term promotes some prior of the signal $\mathbf{x}$,  $\lambda$ is a regularization parameter. 

{We solve} the optimization problem \eqref{eq:optimization_problem} via an  ADMM formulation. To this end, the problem is reformulated as a constrained problem {introducing} the auxiliary variable $\mathbf{z} \in \mathbb{R}^N$

\begin{equation}
    \underset{\mathbf{x},\mathbf{z}}{\text{minimize }} l(\mathbf{x}) + g(\mathbf{z}) \quad \text{subject to} \quad \mathbf{x} = \mathbf{z}, 
\end{equation}

where $g(\mathbf{z})=\lambda {R(\mathbf{z})}$ and $l(\mathbf{x})=\frac{1}{2}\|\mathbf{y}-\mathbf{H_\Phi} \mathbf{x}\|_2^2$. Then the augmented Lagrangian is given by

\begin{equation}
    \begin{aligned}
    L_{p} (\mathbf{x},\mathbf{z},\mathbf{u}) & = \frac{1}{2} ||\mathbf{y} - \mathbf{H_\Phi x}||_{2}^2 + \lambda {R(\mathbf{z})}\\
    & + \frac{\rho}{2} ||\mathbf{u} + \mathbf{x} - \mathbf{z} ||_{2}^2,
    \end{aligned}
\end{equation}
where $\rho>0$ is a penalty parameter {and $\mathbf{u}\in\mathbb{R}^N$} is the dual variable or Lagrange multiplier. The problem can be solved efficiently using ADMM, resulting in the following iterative process:

\begin{equation}
\begin{cases}
   
    \mathbf{z}^{k+1} &:=  \arg \min _{\mathbf{z}} \left( g(\mathbf{z}) + \frac{\rho}{2} \left\| \mathbf{x}^k - \mathbf{z} + \mathbf{u}^k \right\|_2^2 \right), \\
     \mathbf{x}^{k+1} &:= \arg \min _{\mathbf{x}} \left( l(\mathbf{x}) + \frac{\rho}{2} \left\| \mathbf{x} - \mathbf{z}^{k+1} + \mathbf{u}^k \right\|_2^2 \right), \\
    \mathbf{u}^{k+1} &:= \mathbf{u}^k + \mathbf{x}^{k+1} - \mathbf{z}^{k+1},
\end{cases}
\label{eq:iterative_way}
\end{equation}

    {for $k=1,\dots, L$ where $L$ is the total number of iterations {or stages}.} The minimization with respect $\mathbf{z}$ {is a proximal operator over the function $R(\mathbf{z})$. In the unrolling approach, instead of explicitly defining a regularization function, a neural network is employed to learn this proximal mapping. Thus,  employing the neural network $\mathcal{D}_{\omega^{k+1}}(\cdot)$, where $\omega^k$ is the corresponding learnable parameters at each iteration, we have}

\begin{equation}
\mathbf{z}^{k+1}:=\mathcal{D}_{\omega^{k+1}}\left(\mathbf{x}^{k}+\mathbf{u}^k\right).
\end{equation}

For the network structure, we employed the proposed in \cite{ISTA_NET} where an autoencoder is used in the unrolling version of the well-known iterative soft thresholding algorithm (ISTA) for compressed sensing \cite{ISTA_NET}, promoting sparsity in autoencoder latent space. This network is formulated as

\begin{equation}
  \mathcal{D}_{\omega^{k+1}}(\mathbf{x}^{k}+\mathbf{u}^k) = \Tilde{\mathcal{F}}^{k+1}\left( 
 \operatorname{soft}\left( \mathcal{F}^{k+1}(\mathbf{x}^{k}+\mathbf{u}^k), \mathbf{\beta}^{k+1} \right)\right),
\end{equation}

where $\operatorname{soft}(\cdot, \beta^{k+1})$ is the shrinkage thresholding operator, which is parameterized by the learnable parameter $\beta^{k+1}$. {$\mathcal{F}^k$} is a  nonlinear transform that promotes sparsity, and {$\Tilde{\mathcal{F}}^k$} serves as the left inverse operator of {$\mathcal{F}^k$}.  Both {$\mathcal{F}^k$} and {$\Tilde{\mathcal{F}}^k$} are learnable and are modeled as convolution operators followed by a ReLU operator. Thus the trainable parameters of the network to perform the proximal operator are {$\omega^{k} = \{\beta^k,\mathcal{F}^k,\Tilde{\mathcal{F}^k}\}$}. Then, solving $\mathbf{x^{k+1}}$ by gradient descent at each iteration is given by
    \begin{align}
        \mathbf{x}^{k+1} &:= \mathbf{x}^k - \alpha^{k+1} \left( \mathbf{H_\Phi}^T \left( \mathbf{H_\Phi} \mathbf{x}^k - \mathbf{y} \right) \right. \nonumber \\
    & \quad + \rho^{k+1} \left( \mathbf{x}^k - \mathbf{z}^{k+1} + \mathbf{u}^k \right) \Big),
    \end{align}

    where $\alpha$ is the gradient descent parameter. {Here, both $\alpha^k$ and $\rho^k$ are learnable parameters, therefore, the trainable parameters of the recovery network $\psi_\theta(\cdot)$ are $\theta = \{\omega^1,\alpha^1,\rho^1,\dots,\omega^L,\alpha^L,\rho^L\}$. { As initial variables values we used the re-projection of the measurements to the image size \cite{E2E_JORGE_OPTICA}  $\mathbf{x}^0 = \mathbf{H_\Phi}^T\mathbf{H_\Phi x}$ and $\mathbf{u}^{0} = \mathbf{0}$ where $\mathbf{0}$ is a vector full of zeros of an appropriate size.}






\begin{figure*}
    \centering
    \includegraphics[width=0.92\textwidth]{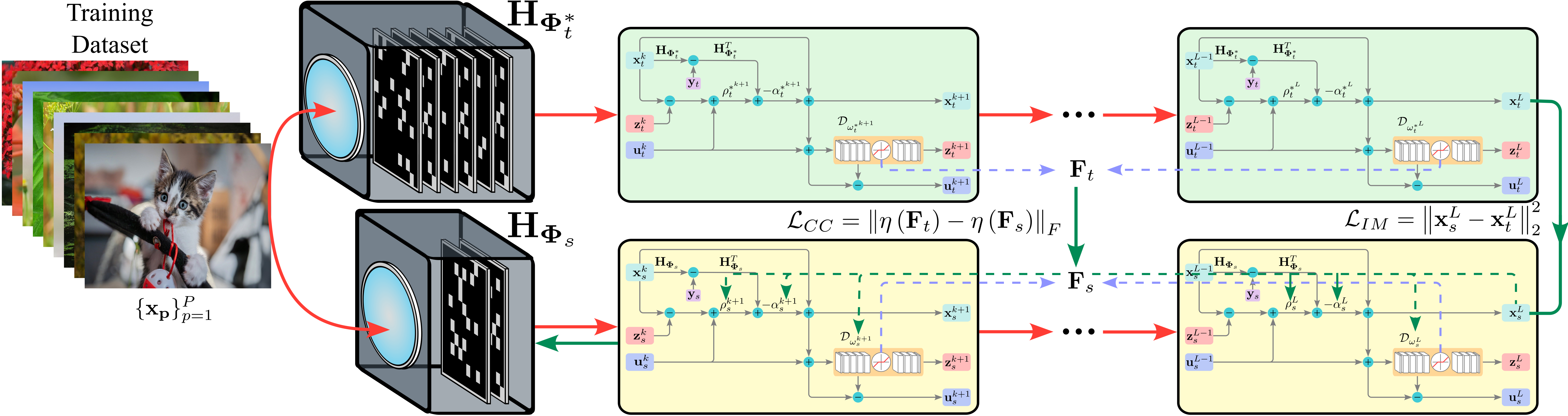}\vspace{-0.4cm}
    \caption{Proposed KD framework, the yellow color is associated with the high-constrained COI system model (student), and the green color is associated with the high-performance, low-constrained COI system model (teacher). \vspace{-0.4cm}}
    \label{fig:KD}
\end{figure*}

\subsection{E2E with Knowledge Distillation } \label{KD}

{In Fig. \ref{fig:KD}, we illustrate the proposed KD approach for designing a highly constrained COI system}. Specifically, a high-performance, low-constrained COI system optimized through E2E, denoted as $\psi_{\theta_t^*}$, with {a sensing matrix} $\mathbf{H_{\Phi_t^*}} \in \mathbb{R}^{M_t \times N}$ {, and OCEs $\mathbf{\Phi_t^*}$  not constrained by $\mathcal{E}$}, is used to transfer its knowledge to a highly constrained COI system, denoted as $\psi_{\theta_s}$ with {a sensing matrix} $\mathbf{H_{\Phi_s}} \in \mathbb{R}^{M_s \times N}$, { where $\mathbf{\Phi_s} \in \mathcal{E}$ and }$M_s \ll M_t$. {The subscript $t$ corresponds to the teacher network, while $s$ corresponds to the student network. To perform KD}, an imitation loss} {\eqref{eq:imitation}} is employed alongside a correlation congruence loss {\eqref{eq:congruenceloss}} \cite{KD_LOSS_CORRELATION}. The imitation loss  compels the student's output $\psi_{\theta_s}(\mathbf{H_{\Phi_s}}^T \mathbf{H_{\Phi_s}} \mathbf{x})$ to mimic the output of the teacher $\psi_{\theta_t^*}(\mathbf{H_{\Phi_t^*}}^T \mathbf{H_{\Phi_t^*}} \mathbf{x})$ . On the other hand, the correlation congruence loss evaluates the correlation congruence between the sparse representation of student and teacher features at all stages of the unrolling recovery network. Let $\mathbf{F}_s$ and $\mathbf{F}_t$ represent the matrices containing the sparse representations of the student and teacher features at every stage of the unrolling recovery network $\mathbf{F}_t = \left[\mathbf{f}^1_t, \mathbf{f}^2_t, \ldots, \mathbf{f}^L_t\right]$and $
        \mathbf{F}_s = \left[\mathbf{f}^1_s, \mathbf{f}^2_s, \ldots, \mathbf{f}^L_s\right],$
 where $\mathbf{f}^k_t \in \mathbb{R}^{NC}$ and $\mathbf{f}^k_s \in \mathbb{R}^{NC}$  represent the sparse representations of the teacher and student features at stage $k$. The term $C$ represents the number of channels of the convolutional layers of the $\mathcal{F}$ operator. The representations  $\mathbf{f}^k_t$ and  $\mathbf{f}^k_s$ are obtained through the $\operatorname{soft}(\cdot, \beta^{k+1})$ shrinkage thresholding operator, with {$\mathbf{f}_t^k = \operatorname{soft}\left( \mathcal{F}_t^{k+1}(\mathbf{x}_t^{k} + \mathbf{u}_t^k), \beta_t^{k+1} \right)$, and $\mathbf{f}_s^k = \operatorname{soft}\left( \mathcal{F}_s^{k+1}(\mathbf{x}_s^{k} + \mathbf{u}_s^k), \beta_s^{k+1} \right)$}.  To capture the complex correlation between the sparse representation features at all stages the {Gaussian radial basis function kernel} is employed. This function measures the similarity between two instances, with a value of 1 indicating closeness and 0 indicating dissimilarity. This correlation function is defined as 
\begin{equation}    {\eta\left(\mathbf{F}\right)=[k(\mathbf{F}, \mathbf{F})]_{i, j}=\exp \left(- \frac{\left\| \mathbf{f}^i-\mathbf{f}^j\right\|_2^2}{2\sigma^2}\right)} ,
\end{equation}
where each element of the correlation matrix $[k(\mathbf{F}, \mathbf{F})]_{i, j}$ encodes the pairwise correlations between {the $\mathbf{f}^i$ and $\mathbf{f}^j$ features} \cite{KD_LOSS_CORRELATION},  $\sigma^2$ represents the variance of the Gaussian distribution, \cite{RBF_KERNEL}, and is a hyper-parameter to be tuned, {a large variance implies that features that are farther apart will have a high similarity value, and a small variance the opposite \cite{RBF_KERNEL_SIGMA}}. The correlation congruence loss is defined as the Frobenius norm of the difference between the correlation matrix of the sparse representation features of the teacher and student network at all stages.
\begin{equation}
\label{eq:congruenceloss}
    \mathscr{L}_{C C}=\left\|\eta\left(\mathbf{F}_t\right)-\eta\left(\mathbf{F}_s\right)\right\|_F .
\end{equation}
Subsequently, the imitation loss is formulated as
\begin{equation}
\label{eq:imitation}
    \mathscr{L}_{IM}=\left\|\psi_{\theta_s}(\mathbf{H_{\Phi_s}}^T \mathbf{H_{\Phi_s}} \mathbf{x})-\psi_{\theta_t^*}(\mathbf{H_{\Phi_t^*}}^T \mathbf{H_{\Phi_t^*}} \mathbf{x})\right\|_2^2 ,
\end{equation}
where $\psi_{\theta_s}(\mathbf{H_{\Phi_s}}^T \mathbf{H_{\Phi_s}} \mathbf{x}) = \mathbf{x}_s^L$ and $\psi_{\theta_t^*}(\mathbf{H_{\Phi_t^*}}^T \mathbf{H_{\Phi_t^*}} \mathbf{x}) = \mathbf{x}_t^L$ represent the recovery of the re-projected measurements of the student and teacher networks at the last stage of the unrolling recovery network (stage $L$).  Then, the optimization goal for KD is to minimize both the imitation loss and the correlation congruence loss as follows \vspace{-0.3cm}
{
\begin{equation}
\label{eq:kd_optimization}
\begin{aligned}
 \{\boldsymbol{\theta}_s^*, \mathbf{\Phi}_s^*\} 
     &=\underset{\substack{ \boldsymbol{\theta}_s, \mathbf{\Phi}}_s}{\arg \min } \ \frac{1}{P} \sum_{p=1}^P \left\|\eta\left({\mathbf{F}_t}_p\right)-\eta\left({\mathbf{F}_s}_p\right)\right\|_F \\ + &  \left\|\psi_{\theta_s}(\mathbf{H}_{\Phi_s}^T \mathbf{H_{\Phi_s}} \mathbf{x}_p)-\psi_{\theta_t^*}(\mathbf{H}_{\Phi_t^*}^T \mathbf{H_{\Phi_t^*}} \mathbf{x}_p)\right\|_2^2, 
\end{aligned}
\end{equation}}
where ${\mathbf{F}_t}_p$ and ${\mathbf{F}_s}_p$ are the feature matrix of the teacher and student model of the $p$-th image of the dataset.
This optimization ensures that not only the student can generate outputs similar to those of the teacher, but also that the congruence between the matrices of the sparse features representation of the student and the teacher is maintained{. Consequently, the student model follows the same feature dynamics across the unrolling network stages as the teacher model, ensuring} an efficient transfer of knowledge from the high-performance, low-constrained model to the high-constrained model.

{\section{Single Pixel Imaging}} \label{sensing_model}
{

To validate the proposed KD framework for designing OCEs of highly constrained COI systems, the SPC for {monochromatic and multispectral image reconstruction was employed}. The architecture of the SPC involves an objective lens that forms an image of the scene onto a binary CA, which encodes the image by either letting pass or blocking the incoming light. {Then}, a collimator lens condenses the encoded image by projecting it onto a single spatial point, where a sensor captures the incoming light intensity \cite{SPC_JORGE, SPC_HANS}. To acquire multiple projections of the same image the CA changes in each acquisition.


Following \eqref{eq:sensing_model}, the SPC  imaging can be modeled as a linear system where all the pixels of an image are mapped to a single pixel. Here $\mathbf{x} \in \mathbb{R}^{MN}$ is the vectorization of the scene with spatial dimensions $M$ and $N$, { $\mathbf{H_\Phi} \in \{-1, 1\}^{K\times MN}$ is the sensing matrix whose rows are the vectorization of the binary CAs $\boldsymbol{\Phi}$. The measurements for $K$ projections is denoted as $\mathbf{y} \in \mathbb{R}^{K}$, and $\boldsymbol{\omega} \in \mathbb{R}^{K}$ is additive noise}. The compression ratio of the SPC is given by the relationship between the number of projections $K$ and the dimension of the given image and is defined as $\gamma=\frac{K}{MN}$, where $\gamma \in [0, 1]$. 

{

For spectral imaging, the sensing model for all snapshots acquired for the $j$-th spectral band can be expressed as

 \begin{equation}
\label{eq:21}
    \mathbf{y}_j = \mathbf{H_\Phi} \mathbf{x}_j + \boldsymbol{\omega}_j,
\end{equation}

where $j={1, \dots, J}$ indexes the spectral bands, $\mathbf{x}_j \in \mathbb{R}^{MN}$ represents the vectorization of the {$j$-th} spectral band of the scene, $\mathbf{y}_j \in \mathbb{R}^{K}$ denotes the measurements, and $\boldsymbol{\omega}_j \in \mathbb{R}^{K}$ is additive noise. In general, for all spectral bands, the sensing model is given by $ \mathbf{\hat{y}} = \mathbf{\hat{H}_\Phi} \mathbf{{x}
    } + \boldsymbol{\hat{\omega}}$, {where $\mathbf{\hat{y}} = \left[\mathbf{y}_1^T, \dots, \mathbf{y}_J^T \right] \in \mathbb{R}^{KJ}$ represents the measurements for all spectral bands, $\mathbf{x} \in \mathbb{R}^{MNJ}$ is the spectral scene, and  $\mathbf{\hat{H}_\Phi} \in \{-1,1\}^{KJ\times MNJ}$ is a block diagonal matrix {defined as} the following {structure}}

\begin{equation}
    \mathbf{\hat{H}_\Phi} = \left[\begin{smallmatrix}
    \mathbf{H_\Phi} & \mathbf{0} & \dots & \mathbf{0}\\
    \mathbf{0} & \mathbf{H_\Phi} & \dots & \mathbf{0} \\
    \vdots & \vdots & \ddots & \vdots \\
    \mathbf{0} & \mathbf{0} & \dots & \mathbf{H_\Phi}
    \end{smallmatrix}\right].
\end{equation}

{ Notice the compression is applied on the spatial domain, such that $K\ll MN$.}
\section{Simulations and Results}

    The {performance} of the proposed KD framework was evaluated in {monochromatic image reconstruction ($J=1$) and multispectral image reconstruction ($J=8$) }tasks using an SPC as the COI system.{ The SPC teacher model considered both real-valued CAs, $\mathbf{H_{\Phi_t^*}} \in \mathbb{R}^{K_t J\times MNJ}$, and binary CAs, $\mathbf{H_{\Phi_t^*}} \in \{-1, 1\}^{K_t J\times MNJ}$, while the SPC student model used binary CAs, $\mathbf{H_{\Phi_s}} \in \{-1, 1\}^{K_s J\times MNJ}$, with $K_s \ll K_t$}. To promote binary CAs, the activation function proposed by \cite{binarized} was employed. This function uses the $\operatorname{sign}$ function in the forward pass, and due to the gradient of this function being zero, in the backward pass the identity function is used, allowing the incoming gradient to substitute the gradient of the $\operatorname{sign}$ function. The reconstruction unrolling network comprises seven iterations or stages. The autoencoder network comprises convolutional layers, followed by a ReLU activation function. The encoder consists of four convolutional layers, each with a kernel size of $3 \times 3 \times C$ {, where $C$ represents the kernel's number of channels, set to $32$}. In the autoencoder's bottleneck, the shrinkage operator is applied. The decoder comprises three convolutional layers, each with a kernel size of $3 \times 3 \times C$. Finally, a convolutional layer is employed to restore the input of the network to its original size, utilizing a kernel size of {$3 \times 3 \times J$}. The FashionMNIST dataset \cite{fashionmnist} was used for the {monochromatic} image reconstruction. This dataset consists of $60,000$ training images of $28\times 28$ with $10$ classes of clothing, divided into $50,000$ for training and $10,000$ for {validation. The test set contains $10,000$ images.} {All} images were resized to $32\times 32$ {pixels}. The ARAD\_1K dataset \cite{ARA1K}  was used for the multispectral image reconstruction task. The images were resized to a size of $100\times 100 \times 8$ extracting $4$ non-overlapping patches per image, obtaining a total of $3600$ patches for training, $125$ for validation, and $75$ for testing. The source code was developed with the PyTorch library and it is available in~\cite{github}.
\subsection{Monochromatic image reconstruction}
In the {monochromatic} image reconstruction experiment, two experiments were conducted to find the best configuration of the teacher network and the best representation of knowledge to be distilled. All teacher and baseline models considered in these experiments were trained for 50 epochs with a learning rate of {$1\times 10^{-4}$} and a batch size of $32$. For all students,  KD was carried out over $50$ epochs with {$\frac{1}{2\sigma^2}=1\times 10^{-6}$,} a learning rate of {$1\times 10^{-3}$} and a batch size of $32$. {The baseline models share the same architecture as the student models. However, instead of being trained through KD, they are trained by E2E optimization following \eqref{eq:e2e_optimization}.}

\begin{figure}[t!]
    \centering
    \includegraphics[width=0.99\columnwidth]{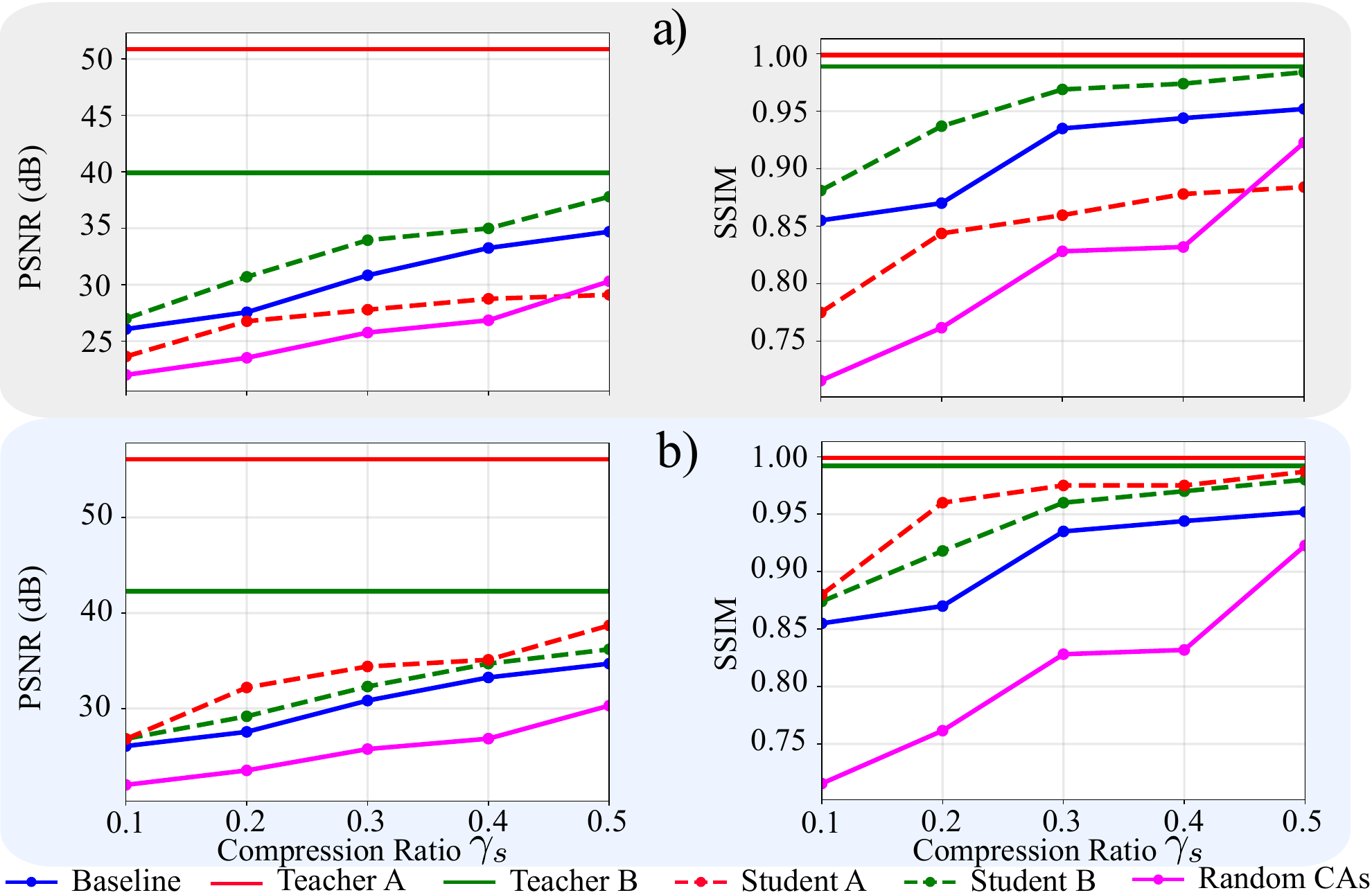}
    \caption{{Comparison of the reconstruction performance among the E2E baseline, the student models trained without optimizing their OCEs, i.e., employing random binary CAs, and the KD approach. a) Teachers with $\gamma_t=0.8$ and b) Teachers with $\gamma_t=0.9$. Teacher A is a real-valued CA SPC, Teacher B is a binary CA SPC, and Students A and B are binary CA SPCs distilled from teachers A and B, respectively.}}\vspace{-0.6cm}
    \label{fig:spc_recovery}
\end{figure}
\subsubsection{Optimal teacher configuration}
\begin{figure*}[t!]
    \centering
    \includegraphics[width=0.85\textwidth]{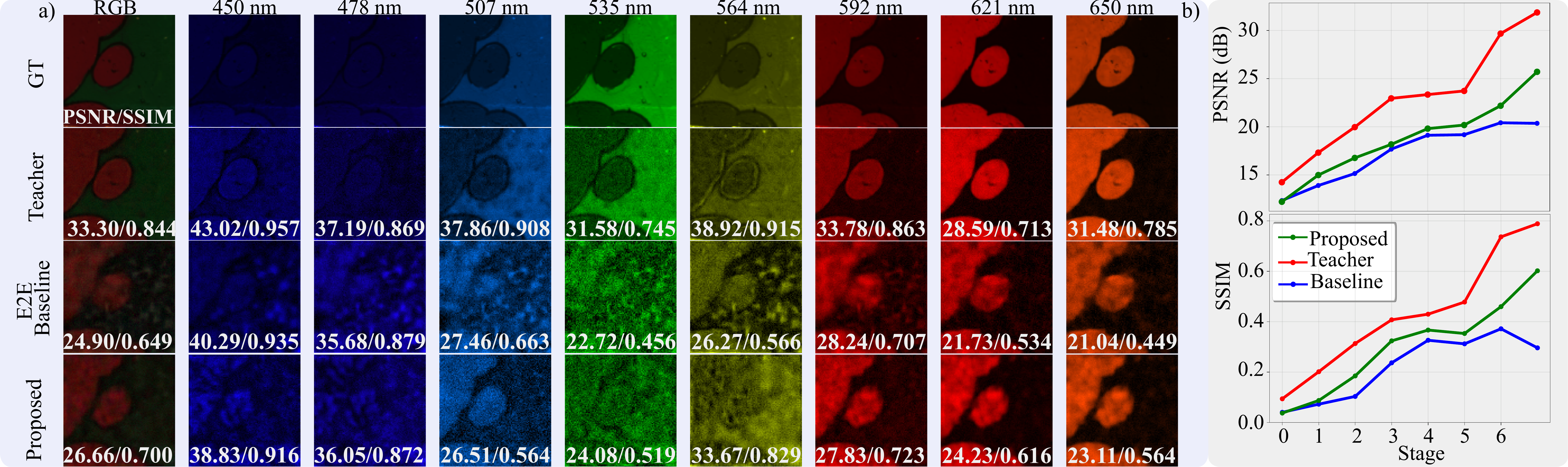}\vspace{-0.3cm}
    \caption{{Comparison of the reconstruction performance among the teacher model, E2E baseline, and the proposed KD approach for reconstructing eight spectral bands. a) Displays the RGB representation of the reconstructed SI of the baseline and the proposed method. Also, the eight bands are plotted. b) shows the recovery performance of the test dataset along the stages.}\vspace{-0.6cm}  }
    \label{fig:spectral}
\end{figure*}
In this experiment, KD was performed from different high-performance, low-constrained COI systems (the teacher model). {These teacher models considered two types of constraints: the number of snapshots, and the values that the CAs $\boldsymbol{\Phi}$ can take, either binary or real-valued. Consequently, our teacher models comprised a binary CA SPC with $\gamma_t=0.8$, a binary CA SPC with $\gamma_t=0.9$, a real-valued CA SPC with $\gamma_t=0.8$, and a real-valued CA SPC with $\gamma_t=0.9$.} 

\begin{table}[!b]
 \vspace{-0.3cm}\caption{Comparison of recovery performance among the E2E baseline, KD with sparse features, and KD with non-sparse features. The best results are highlighted in green, while the second-best results are presented in blue.}

\label{tab:sparse-no-sparse}
\resizebox{\columnwidth}{!}{%
\footnotesize
\begin{tabular}{ccccccc}

\hline
\multirow{2}{*}{$\gamma_s$} & \multicolumn{2}{c}{Baseline} & \multicolumn{2}{c}{Sparse Distillation} & \multicolumn{2}{c}{No Sparse Distillation} \\ \cline{2-7} 
                    & PSNR (dB)       & SSIM       & PSNR (dB)            & SSIM             & PSNR (dB)              & SSIM              \\ \hline
0.1                 & 26.07          & 0.86      & \cellcolor{blue!12}26.99               & \cellcolor{blue!12}0.88            & \cellcolor{green!12}27.10                 & \cellcolor{green!12}0.89             \\
0.2                 & 27.56          & 0.87      & \cellcolor{green!12}30.70               & \cellcolor{green!12}0.94            &\cellcolor{blue!12} 28.85                 & \cellcolor{blue!12}0.91             \\
0.3                 & \cellcolor{blue!12}30.83          & \cellcolor{blue!12}0.94      & \cellcolor{green!12}33.95               & \cellcolor{green!12}0.97            & 29.42                 & 0.90             \\
0.4                 & \cellcolor{blue!12}33.25          & \cellcolor{blue!12}0.94      & \cellcolor{green!12}34.99               & \cellcolor{green!12}0.97            & 32.45                 & 0.94             \\
0.5                 & 34.70          & 0.95      & \cellcolor{green!12}37.80               & \cellcolor{green!12}0.98            & \cellcolor{blue!12}37.10                 & \cellcolor{blue!12}0.98             \\ \hline
\end{tabular}%
}
\end{table}

After training the four teacher models, each was employed to distill its knowledge into a set of highly constrained COI systems (the student models).  Specifically, five student models with different {$\gamma_s$ were utilized, each consisting of a binary CA SPC with $\gamma_s \in \{0.1,0.2,0.3,0.4,0.5\}$}. {KD of the sparse representation of the teacher features was conducted, and the students were compared against the baseline models, and the same model as the students but without optimizing their OCEs, i.e., employing random binary CAs}. Qualitative evaluation of these experiments utilized the average value of the PSNR and SSIM metrics. The obtained results are presented in Figure \ref{fig:spc_recovery}, { where the five student networks distilled from the real-valued CA SPC teacher with $\gamma_t=0.8$ showed low performance, those distilled from the other three teachers achieved high performance, surpassing the baseline models optimized via E2E, {and the student models trained without optimizing their OCEs}. Nevertheless, it can be noted that despite the high performance achieved by these students, a significant performance gap exists between the teacher and its students, particularly with the real-valued CA SPCs with $\gamma_t=0.9$ and binary CA SPCs with $\gamma_t=0.9$ teachers. The binary CA SPC teacher with $\gamma_t=0.8$ demonstrated the best ratio between its performance and that of its students, despite not being the best-performing teacher. This }can be attributed to the {different nature} between the teacher and student models. Past studies \cite{KD_GAP_1, KD_GAP_2} have shown that when there is a notable difference in complexity, the performance of the student model tends to drop. Notably, \cite{KD_GAP_1} has found that students distilled from high-performance teachers might perform worse than those distilled from a more compact teacher.


\subsubsection{Optimal representation of knowledge}
This experiment aims to determine the optimal representation of knowledge to be distilled using the correlation congruence loss. Two types of features were considered as knowledge. The first type involves the sparse representations of the teacher features, corresponding to the output of the shrinkage thresholding operator of the autoencoder network at all stages of the recovery network. The second type considers the non-sparse representations of teacher features, corresponding to the output at every stage of the recovery network. This experiment considers the optimal teacher identified in the previous experiment: the binary CA SPC with $\gamma_t=0.8$. Both types of knowledge representations were distilled from the teacher to the five student models considered in the previous experiment. {The performance of students trained with these knowledge representations was compared to the baseline models}. Performance evaluation employed the {average of the }PSNR and SSIM metrics. Table \ref{tab:sparse-no-sparse} shows the obtained results, where the best representation of knowledge was the sparse features. This {can be related to} the sparse features often provide a more compact representation of information, focusing on essential elements and reducing redundancy. 
 \subsection{Multispectral image reconstruction }
This experiment is aimed at validating the proposed framework in a more challenging scenario, where multispectral image reconstruction takes place. The teacher network consisted of a binary CA SPC with {$\gamma_t=0.6$ {, and was} trained for $50$ epochs with a batch size of $4$ and a learning rate of {$1\times 10^{-3}$}. The student model consisted of a binary CA SPC with $\gamma_s=0.1$}. {The} KD of the sparse representations of the teacher features was carried out over 25 epochs with a batch size of $4$, a learning rate of {$1\times10^{-3}$}, and {$\frac{1}{2\sigma^2}=1\times10^{-6}$.} The student was compared against the same model trained by E2E optimization following \eqref{E2E_optimization} (the baseline model). Qualitative evaluations of the performance of the models include the {average of the} PSNR and SSIM metrics. The obtained results are presented in Fig. \ref{fig:spectral}. These results demonstrate a performance improvement obtained by distilling the knowledge from the teacher compared to the E2E optimization of the baseline model.\vspace{-0.2cm}

\section{Conclusion and Future Outlooks} \label{conclusion}
{A KD framework for optimizing OCEs in highly {physically} constrained COI systems was proposed. {To address} the limitations {presented} by the highly constrained E2E optimization of OCEs in {these} systems, {this framework} transfers the knowledge {from} a high-performance, low-constrained COI system {(the teacher)} to a highly constrained COI system { (the student)}. It was found that the best knowledge representation was the sparse teacher features, as they may provide a compact representation of knowledge reducing redundancy. Furthermore, it was validated that the best teacher, is not always the best performing, showing that a more compact teacher performed better to those high performance. The proposed framework was validated in two recovery tasks employing the SPC. Simulations show an improvement in the design of OCEs of highly physically constrained COI systems through KD. { Future research can extend KD to other OCEs, such as diffractive optical elements and color filter arrays, and to other computational imaging systems, such as computed tomography and seismic imaging.} }

\newpage\newpage
\footnotesize{\bibliographystyle{IEEEbib}}
\bibliography{strings,refs}

\end{document}